\documentclass[conference]{IEEEtran}
\IEEEoverridecommandlockouts
\usepackage{cite}
\usepackage{amsmath,amssymb,amsfonts}
\usepackage{float}
\usepackage{algorithmic}
\usepackage{graphicx}
\usepackage{textcomp}
\usepackage{xcolor}
\usepackage[hidelinks]{hyperref}
\hypersetup{
  colorlinks,
  allcolors=.,
  urlcolor=blue,
}
\def\BibTeX{{\rm B\kern-.05em{\sc i\kern-.025em b}\kern-.08em
    T\kern-.1667em\lower.7ex\hboxhttps://www.overleaf.com/project/6382de28477ed74f829a2fed{E}\kern-.125emX}}
    
\DeclareMathOperator*{\E}{\mathbb{E}}

\begin{document}

\title{Predictive World Models from Real-World Partial Observations\\
}

\author{\IEEEauthorblockN{1\textsuperscript{st} Author}
\IEEEauthorblockA{\textit{Department} \\
\textit{Organization}\\
Location \\
Email}
\and
\IEEEauthorblockN{2\textsuperscript{nd} Author}
\IEEEauthorblockA{\textit{Department} \\
\textit{Organization}\\
Location \\
Email}
\and
\IEEEauthorblockN{3\textsuperscript{rd} Author}
\IEEEauthorblockA{\textit{Department} \\
\textit{Organization}\\
Location \\
Email}
\and
\IEEEauthorblockN{4\textsuperscript{th} Author}
\IEEEauthorblockA{\textit{Department} \\
\textit{Organization}\\
Location \\
Email}
\and
\IEEEauthorblockN{5\textsuperscript{th} Author}
\IEEEauthorblockA{\textit{Department} \\
\textit{Organization}\\
Location \\
Email}
}

 \author{\IEEEauthorblockN{1\textsuperscript{st} Robin Karlsson}
 \IEEEauthorblockA{\textit{Graduate School of Informatics} \\
 \textit{Nagoya University}\\
 Nagoya, Japan \\
 karlsson.robin@g.sp.m.is.nagoya-u.ac.jp}
 \and
 \IEEEauthorblockN{2\textsuperscript{nd} Alexander Carballo}
 \IEEEauthorblockA{\textit{Department of Electrical, Electronic and Computer Engineering} \\
 \textit{Gifu University}\\
 Gifu, Japan \\
 alex@gifu-u.ac.jp}
 \and
 \IEEEauthorblockN{3\textsuperscript{rd} Keisuke Fujii}
 \IEEEauthorblockA{\textit{Graduate School of Informatics} \\
 \textit{Nagoya University}\\
 Nagoya, Japan \\
 fujii@i.nagoya-u.ac.jp}
 \and
 \IEEEauthorblockN{4\textsuperscript{th} Kento Ohtani}
 \IEEEauthorblockA{\textit{Graduate School of Informatics} \\
 \textit{Nagoya University}\\
 Nagoya, Japan \\
 ohtani.kento@g.sp.m.is.nagoya-u.ac.jp}
 \and
 \IEEEauthorblockN{5\textsuperscript{th} Kazuya Takeda}
 \IEEEauthorblockA{\textit{Graduate School of Informatics} \\
 \textit{Nagoya University, TIER IV}\\
 Nagoya, Japan \\
 kazuya.takeda@nagoya-u.jp}
 }

\maketitle
\thispagestyle{plain}
\pagestyle{plain}

\begin{abstract}
Cognitive scientists believe adaptable intelligent agents like humans perform reasoning through learned causal mental simulations of agents and environments. The problem of learning such simulations is called predictive world modeling. Recently, reinforcement learning (RL) agents leveraging world models have achieved SOTA performance in game environments. However, understanding how to apply the world modeling approach in complex real-world environments relevant to mobile robots remains an open question. In this paper, we present a framework for learning a probabilistic predictive world model for real-world road environments. We implement the model using a hierarchical VAE (HVAE) capable of predicting a diverse set of fully observed plausible worlds from accumulated sensor observations. While prior HVAE methods require complete states as ground truth for learning, we present a novel sequential training method to allow HVAEs to learn to predict complete states from partially observed states only. We experimentally demonstrate accurate spatial structure prediction of deterministic regions achieving 96.21 IoU, and close the gap to perfect prediction by 62 \% for stochastic regions using the best prediction. By extending HVAEs to cases where complete ground truth states do not exist, we facilitate continual learning of spatial prediction as a step towards realizing explainable and comprehensive predictive world models for real-world mobile robotics applications. Code is available at \url{https://github.com/robin-karlsson0/predictive-world-models}.
\end{abstract}

\section{Introduction}

\begin{figure}
\centering
\includegraphics[width=0.45\textwidth]{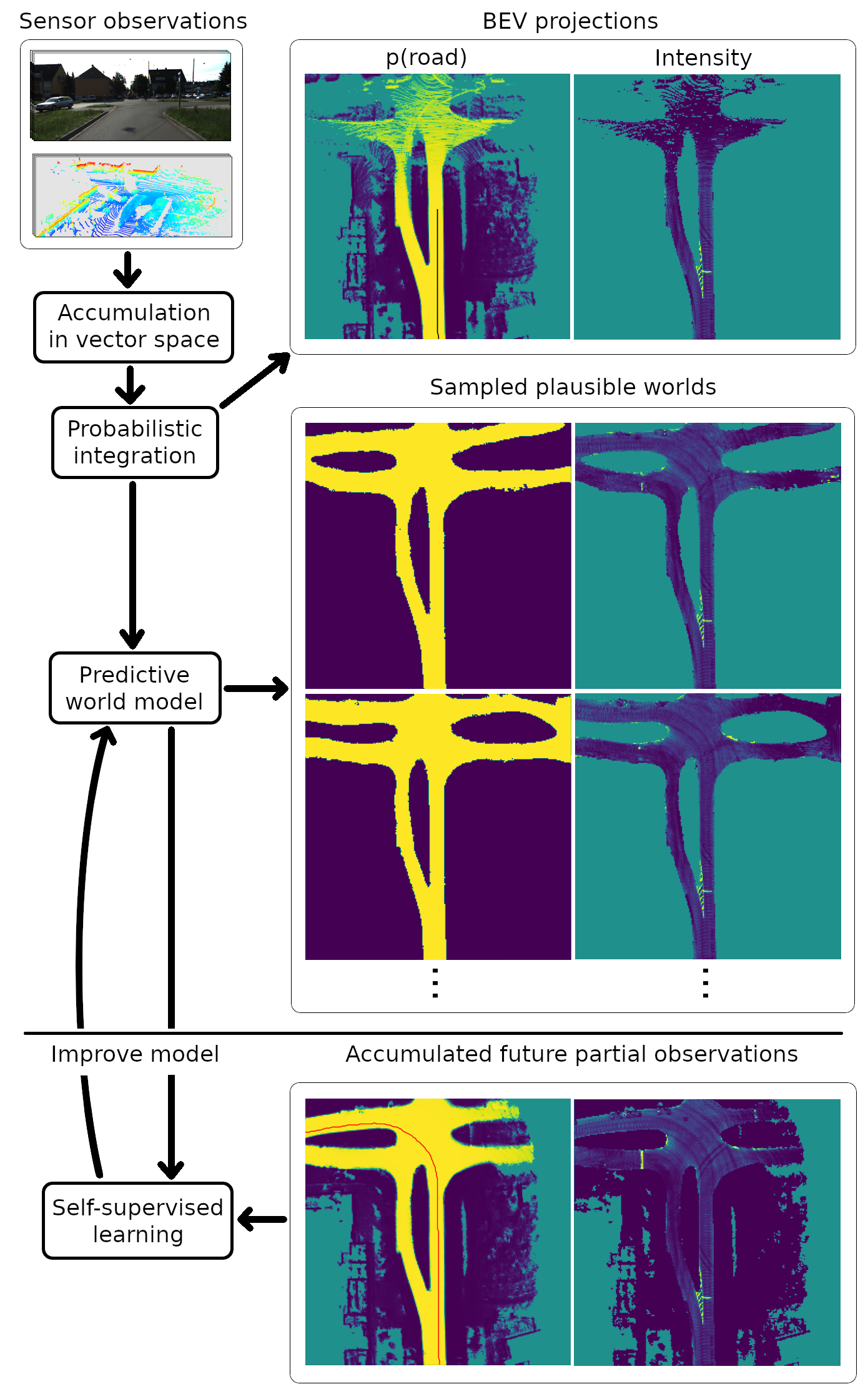}
\caption{The framework integrates observations into a common vector space representing the partially observed world state. A predictive world model samples a set of diverse plausible complete world states. The model improves through continual learning from experience by predicting future observations.}
\label{fig:front_figure}
\end{figure}

Cognitive scientists believe cognition in adaptable intelligent agents like humans is founded on a small number of foundational components for representing the world in terms of inanimate objects, goal-directed agents, number systems and sets, social partners and groups, and geometry of environments~\cite{spelke2007CoreK}. These cognitive abilities allow intelligent agents to perform common-sense physical reasoning to facilitate accomplishment of tasks~\cite{lake2017BuildingMachines}. One ability is to predict multiple plausible long-tail outcomes of a sequence of actions, and perform counterfactual reasoning~\cite{pearl1988ProbabilisticRI, pearl2000Causality, pearl2019SevenTools} through causal mental simulations of the world. Another ability is to imagine different plausible spatial configurations of unobserved regions of the world based on past experience. The problem of learning such simulations is called predictive world modeling in machine learning~\cite{schmidhuber90world-model, schmidhuber91curiosity-boredom, schmidhuber10-formal-theory}.

%
The explicit world modeling approach provides several potential advantages over implicit predictive models learned as part of the task; data efficiency as policies can be optimized through simulation, long-tail planning and reasoning by search, explicit representation of agent's knowledge and state space coverage, explainable sequential decision processes, and improved domain generalization by enabling planning over abstract latent structures decoupled from particular observable appearance (i.e. pixels).
The advantage of world models is demonstrated by recent model-based reinforcement learning (RL) methods like Dreamer V2~\cite{hafner2021MasterAW} demonstrating better performance than SOTA model-free RL methods like IQN~\cite{Dabney2018IQN} and Rainbow~\cite{Hessel2018Rainbow} on Atari environments~\cite{bellemare2013Atari} when compared by the same amount of compute and wall-clock time.
However, understanding how to learn and apply the world modeling approach in visually complex real-world environments relevant to mobile robots, and in particular autonomous vehicles (AV), remains an open question.
Learning world models of real-world environments is associated with challenges.
Agent observations are perceptually complex, making it difficult to decompose observations into consistent and distinct semantic objects, in turn making relationship learning infeasible.
Percepts are generally agent-centric, partial observations due to occlusion and limited sensor observation reach, meaning explicitly inferring the complete world state is difficult if not impossible.
World state transition dynamics resulting from multi-agent interactions are generally stochastic and complex.
Tasks are generally specified vaguely and lack rich reward signals.
In contrast to game environments, experience gathering is radically limited in real-world environments, as failure may have unacceptable consequences and generally prohibit learning world states primarily by exploration.

Mobile robots typically perform motion planning on the assumption that the spatial environment is fully known~\cite{Paden2016MPSurvey, Claussmann2019MPReview}.
This assumption is conventionally satisfied through localizing the agent within point cloud maps encoded with layers of human-annotated HD map information about the environment~\cite{Sheif2016HDMaps}.
However, solely relying on a priori maps has demerits~\cite{Karlsson2021gdsla}. Map generation proves difficult to scale, as creating maps require labor-intensive human annotation and regular updating as environments change. Dependence on maps for navigation renders the agent inoperable in case of loading or localization failure, as well as in unmapped environments. Additionally, changes in the environment not reflected in the map may result in unsafe behavior.

This work proposes a framework for learning probabilistic predictive world models for real-world spatial environments from self-supervised learning using sensor observations. We implement the world model based on the recent SOTA hierarchical VAE (HVAE) model Very Deep VAE (VDVAE)~\cite{Child21VDVAE} capable of representing and sampling general and spatially large real-world environments by a compact latent code while preserving fine spatial detail. We introduce two novel components to overcome the fact that the original VDVAE model cannot learn to predict complete states from partially observed states only. First, we demonstrate how to generate pseudo-complete states using a combination of latent variable predictive and adversarial modeling. Secondly, we present an approach to enable HVAEs to predict a diverse set of complete states from a single partial state by learning to match the latent variable distributions between the partial and pseudo-complete states.

By enabling learning from partial observations generated by sensor information, our predictive world model becomes capable of improving from new experiences obtained during operation akin to continual learning~\cite{thrun1995lifelong_robot, cavallari2017on-the-fly, lesort2020cont_learn_robot}. The model can avoid catastrophic forgetting by retaining a replay buffer of past experiences~\cite{minh2015dqn}.


We consider a structured state description of the spatial world as a discrete, agent-centric, 2.5D homogeneous grid map to represent probabilistic spatiosemantic information of the environment as observed and predicted by the agent. The grid map is considered a well-established and adequate approach for representing spatial information in mobile robotics~\cite{Thrun2005ProbRob} and compatible with potent generative image modeling approaches in machine learning.


We propose several useful mobile robotics applications; 
robust and safe planning by taking into account diverse sampled structures for unobserved regions; 
improve localization matching success through densifying observations and predicting plausible structure of unobserved regions;
and verify map consistency with the actual perceived environment.

The contributions of our paper are fourfold:
\begin{itemize}
    \item A new conceptual approach to predict a diverse set of plausible, fully observed, real-world spatial environments by a predictive world model and sensor observations.
    \item A novel method to train a HVAE to predict complete states from partially observed states only.
    \item A holistic mobile robotics framework tying together real-world sensor observations and world modeling as a self-supervised learning problem.
    \item Demonstrate accurate spatial structure prediction of deterministic regions achieving 96.21~\% IoU with 1 sample, and close the gap to perfect prediction by 62~\% for stochastic regions using the best prediction out of 32 samples.
\end{itemize}

%

\section{Related work}

\subsection{Arbitrary conditional density estimation}

%
%
The problem of arbitrary conditional density estimation\cite{ivanov2019ac_vae, li2020ac_flow, strauss2021ac_energy} is about estimating the probability distributions $p(x_u|x_o)$, where the random variables $x$ are expected to be partitioned into arbitrary plausible subsets of observed $x_o$ and unobserved $x_u$ random variables.
In this section we present methods incorporating different application-specific presumptions on how $x$ is partitioned into $x_o$ and $x_u$.

Image inpainting methods predict unobserved pixels $x_u$ from observed pixels $x_o$. The problem formulation is similar to the problem of predicting complete world states from partially observed states. The prototypical solution is to use an autoencoder (AE)~\cite{ballard1987ae} to compress partially observed images $x_o$ into constrained latent codes $z$ encoding similar visual patterns as learned from reconstructing complete images $x$ by matching global contextual clues. 

However, optimizing models simply by pixel-wise reconstruction is afflicted by the marginalization problem, resulting in blurry outputs as missing regions can be filled by many plausible pixel configurations. The Context encoder~\cite{pathak2016context_encoder} attempts to address the blurriness problem by introducing an adversarial objective.
Furthermore, GLCIG~\cite{iizuka2017glcig} introduces a course-to-fine generation scheme with diluted convolutions and two adversarial objectives. The global objective ensures the image remains coherent as a whole, while the local objective improves detail.
Yeh et al.~\cite{yeh2017inpainting_gen} find the closest sample in an image database and use its latent code for prediction. 
Contextual attention~\cite{yu2018contextual_attention} adds an attention mechanism for long-distance information crossover.
Our framework similarly applies an adversarial objective for learning to predict texture-like content such as lidar reflectance intensity from road surface (henceforth, road surface intensity).

Other approaches focus on learning mask-aware convolutional filters.
Liu et al.~\cite{liu2018partial_convs} introduces a special convolution filter and a observed element mask update rule for propagating information about which elements provide information.  
Yu et al.~\cite{yu2019gated_convs} introduces gated convolutions for learned mask updating.
While we add an observed element mask to the model input following the missing data VAE approach, explicitly convoluting over masks is an interesting future direction.

Another line of image inpainting works focuses on pluralistic stochastic state completion methods based on generative models.
GAN-based methods~\cite{cai2020piigan, liu2021pdgan} generate multiple plausible completions by conditioning on a random vector.
VAE-based methods~\cite{kingma2013} replace the deterministic latent code generated by the AE to allow stochastic sampling of multiple plausible predictions. Previous methods improve training stability by constraining the latent distribution of partially observed images by matching the distribution for fully observed images.
PIC-Net~\cite{zheng2019picnet} trains separate encoders for observable and unobservable image regions and matches the distributions between the two.
UCTGAN~\cite{zhao2020uctgan} adds a cross-attention module to mix latent representations of partially and fully observed images.
DSI-VQVAE~\cite{peng2021dsi-vqvae} applies VQVAE to stabilize training.
Concurrently to our work, Posterior Matching~\cite{Strauss2022PosteriorMatching} presents arbitrary conditioning based on HVAEs by optimizing a secondary partially observed encoder to match the latent distributions of a fully observed encoder.
We extend prior VAE work by introducing a two-stage training paradigm to allow learning to predict complete images from partially observed images only.

Another approach frames predicting unobserved state variables from observed variables as the missing data VAE problem.
HI-VAE~\cite{nazabal2018hi-vae} derives an evidence lower bound (ELBO) for missing data by masking out contributions from unobserved data.
EDDI~\cite{ma2019eddi} introduces an alternative Partial VAE model which processes observable data only by 
encoding elements by a positional encoding and processed by permutation invariant operations similar to PointNet~\cite{qi2017pointnet}.
VAEM~\cite{ma2020vaem} is a hierarchical VAE that operates on heterogeneous data by first transforming all input variables into a common latent space by a type-specific transformation.
HH-VAEM~\cite{peis2022hh-vaem} is a recent hierarchical VAE demonstrating effective sampling using the Hamiltonian Monte Carlo algorithm.
Collier et al.~\cite{collier2020vae_missing} demonstrate results on high-dimensional image data. 
Our work extends prior missing data VAE approaches by learning to model $p(x_u|x_o)$ for high-dimensional representations without requiring fully observed ground truth samples for training.

Video prediction methods aim to model a stochastic state transition process where a sequence of future images $x_u$ are predicted conditioned on a sequence of past fully observed images $x_o$.
Babaeizadeh~\cite{babaeizadeh2018var_video_pred} presents a sequential stochastic variational video prediction model based on predicting a latent code explaining away the stochasticity of the sequence.
Denton~\cite{denton2018sto_vid_pred} presents an end-to-end framework to explain away stochasticity by a frame-to-frame latent code and a learned prior to improve training robustness.
Our work reformulates the stochastic latent variable video prediction approach of Denton to the problem of predicting complete world states from partially observed world states only.

\subsection{Bird's-eye-view generation}
Mobile robotics, and in particular AVs, pursue the problem of generating top-down bird's-eye-view (BEV) representations from perception inputs as a substitute or complement to human annotated maps~\cite{Karlsson2021gdsla}.

Camera-based methods receive much attention because of affordability and motivation by human vision. However, lifting 2D perspective images to 3D is fundamentally an ill-posed problem.
Inverse perspective mapping (IPM)~\cite{mallot1991ipm, bertozzi1998gold_ipm, bertozzi1998stereo_ipm} proposes to overcome the problem by assuming the ground plane is flat. However, the flat plane assumption is generally not true.
Stereo cameras propose to solve the lifting problem by inferring depth maps based on physics. However, the resulting depth maps tend to be noisy for far-away objects, object borders, and objects covered with non-distinct textures.
Learning-based methods are proposed to overcome the weaknesses of stereo-based depth map estimation.
Cam2BEV~\cite{reiher2020cam2bev} presents an approach that projects semantic features using IPM and corrects the projection by a spatial transformer module learned from synthetic ground truth BEVs. 
Many works are based on using monocular depth estimation~\cite{wang2019pseudo-lidar, quian2020ee-pseudo-lidar, you2020pseudo-lidar++, guizilini2020sem_guided_depth, guizilini20203d-packing, guizilini2021san} to lift images to a 3D point cloud before projection to a top-down 2D grid.
Schulter et al.~\cite{schulter2018look_around_objects} proposes an adversarial objective relying on ground truth maps to refine the resulting BEV representation.
MonoLayout~\cite{mani2020monolayout} learns the view transformation from self-supervised targets by integrating projected observations while still relying on ground truth maps for BEV refinement.
Later works introduce probabilistic depth projection~\cite{philion2020lift-splash-shoot}, 
categorical depth distribution network~\cite{reading2021caddn}, and
multi-task learning~\cite{hu2021fiery}.
VED~\cite{lu2018ved} is a variational encoder trained from stereo vision to predict low-dimensional (64x64 px) semantic BEV representations from forward-view monocular images.
Other methods lift images using multilayer perceptrons (MLP) trained on ground truth maps~\cite{roddick2018oft, roddick2020PyrOccNet, hendy2020fishing-net}.
Recently, cross-attention based transformer modules~\cite{yang2021pyva, wang2021detr3d} 
and Transformers modules~\cite{chitta2021neat} are applied to model view transformations motivated by the global attention mechanism not being limited to processing neighboring pixel information like CNNs. However, due to lacking inductive biases attention-based models tend to require more data, effort, and compute to train as well as for inference.
While our framework in principle is compatible with depth estimation, we choose to leverage lidar for substantial improvements in representation accuracy and observation integration performance. Additionally, our generative model can generate diverse plausible predictions, unlike view transformation models which typically are deterministic functions.


Lidar-based BEV generation methods have a significant advantage from explicitly measuring distance though deemed prohibitively expensive for mass deployment by some.
%
%
Fishing Net~\cite{hendy2020fishing-net} utilizes lidar information to improve spatial accuracy of BEVs generated by sensor fusion.
MP3~\cite{casas2021mp3} uses a learned module for generating map elements from lidar observations and ground truth map supervision.
HDMapNet~\cite{li2022hdmapnet} also includes image information.
In contrast to these methods, our framework does not rely on preexisting ground truth maps for supervised training. Our method is also generative and can provide diverse predictions, which is fundamentally necessary as the correct prediction for occluded regions are generally indeterminable.

\subsection{Spatial AI}
Simultaneous localization and mapping (SLAM)~\cite{smith1986slam1, smith1986slam2, thrun2007stanley} is the conventional robotics approach to mapping 3D spatial environments. SLAM works by computing the translation and rotation transformation to optimally match sequential point clouds. Knowing the transformation allows accumulation of point clouds in a common reference frame or vector space. Our framework integrates sensor observations using the same principle. Another component of SLAM is loop closure optimization when previously traversed spaces are revisited. Our framework can be considered as adding a predictive component to SLAM.

Neural Radiance Fields (NeRF)~\cite{mildenhall2020nerf} is a recent approach to represent 3D objects~\cite{ost2021neural_scene_graph, mildenhall2022nerf_dark} and environments~\cite{brualla2021nerf_wild, rematas2022urban_rad_field} by neural networks. While NeRFs can interpolate between observations they do not extrapolate beyond what is observed like our framework.

\subsection{World models}
The idea of learning a predictive model of the world in machine learning was first introduced by Schmidhuber~\cite{schmidhuber90world-model, schmidhuber91curiosity-boredom, schmidhuber10-formal-theory}.
%
%
A common approach is to learn latent state representations from images using a VAE~\cite{corneil2018vast, ha2018wm, corneil2018tabular_wm}, and use the learned latent code as a compact representation of the world state for planning actions. Other works use adversarial learning to optimize the latent code~\cite{kurutach2018causal_infogan, wang2019viz_plan_act}, or contrastive learning with latent variables to model stochastic transition processes~\cite{lecun2022jepa}.

Another line of work focuses on inferring a set of object encodings from images. 
Watters et al.~\cite{watters2017vin} uses a variational encoder to infer a fixed set of latent object encoding vectors from a sequence of images.
Later works apply a VAEs to learn semantically richer object embeddings~\cite{hafner2019latdyn, laversanne2018cur_goal_spaces}.
MONet~\cite{burgess2019monet} is a prominent model for learning to extract a variable amount of semantic object encodings using a recurrent attention module.
Recent works leveraging MONet demonstrate the merits of explicit object discovery for future state prediction using compositional reasoning~\cite{kipf2020cl_wm}, 
and for reinforcement learning~\cite{watters2019cobra, hafner2021MasterAW} surpassing the performance of SOTA model-free RL models~\cite{Dabney2018IQN, Hessel2018Rainbow}.

Our work approaches the world modeling problem of learning to predict a 2D spatiosemantic representation from agent-centric partial observations. We hope our method will contribute towards bridging recent SOTA world modeling approaches from game environments to partially observed real-world mobile robotics environments.

\section{Generating partial world states}
\label{sec:generating_partial_world_states}
\begin{figure*}[htbp]
\centering
\includegraphics[width=\textwidth]{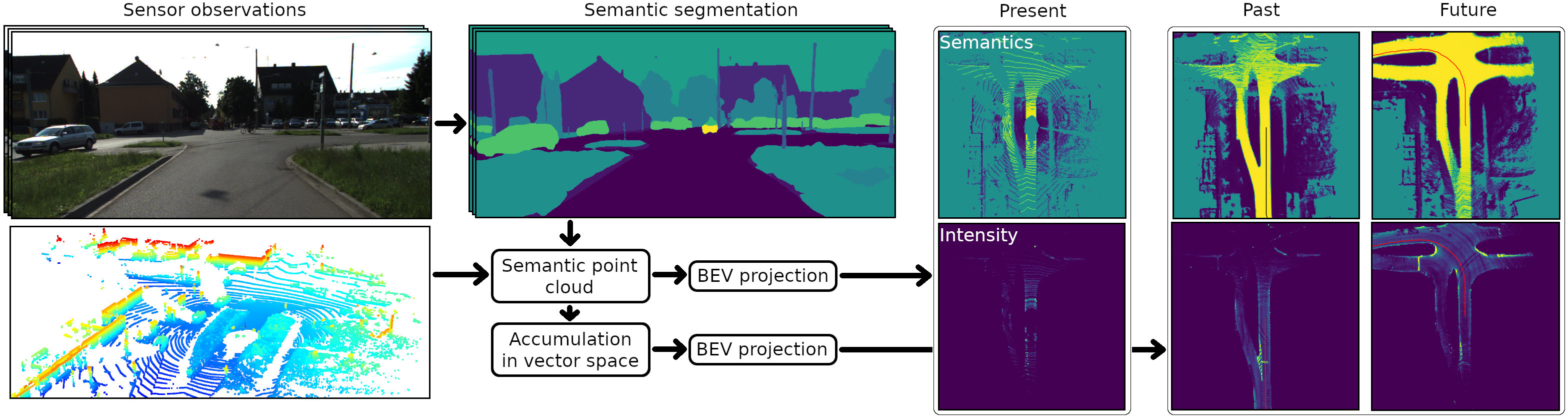}
\caption{Perception pipeline for transforming synchronized sensor observations into BEVs by projection into a common vector space. A semantic segmentation model interprets images. The inferred semantics are attached to the point clouds. Multiple semantic point clouds are temporally integrated into an ego-centric reference frame. BEVs are generated by projecting and probabilistically integrating all semantic points into a 2D discretized grid.}
\label{fig:observation_processing}
\end{figure*}

This section describes the process of turning a sequence of sensor observations into partially observed world states. We conceptualize our approach as the most elementary and general way of achieving spatiosemantic cognition based on projecting and integrating observations into a common metric vector space. See Fig.~\ref{fig:observation_processing} for a visual overview.

\begin{figure}[htbp]
\centering
\includegraphics[width=0.48\textwidth]{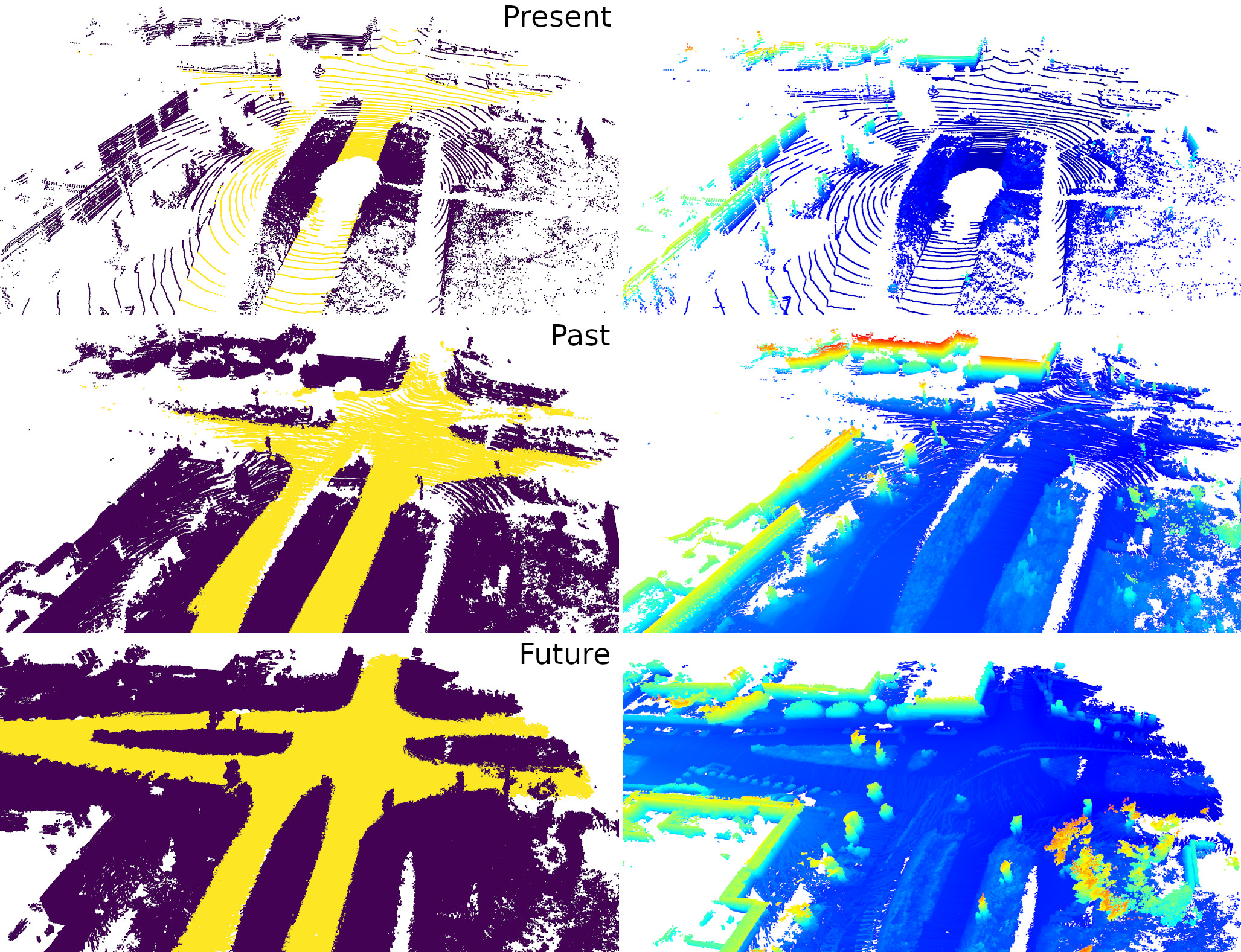}
\caption{Temporal accumulation turns sparse observations (i.e. present) into dense representations (i.e. past and future). Partitioning the accumulated semantic point cloud in past and future observation subsets provide a natural self-supervised learning signal. The left figures show ``road'' (yellow) and ``not-road'' (purple) semantics. The right figures show height information.}
\label{fig:semantic_point_cloud_representation}
\end{figure}

\begin{figure}
\centering
\includegraphics[width=0.48\textwidth]{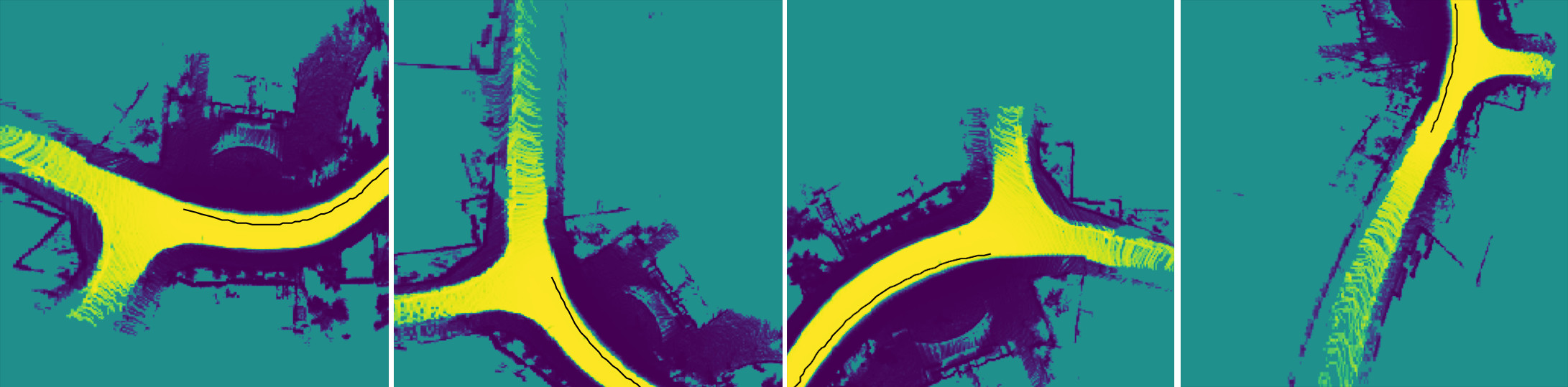}
\caption{Examples of a geometrically diverse set of training samples generated from a single set of real observations using data augmentation.}
\label{fig:data_augmentation}
\end{figure}

\subsection{Sensor observation processing}
\label{sec:sensor_observations}
Perception sensor configurations for mobile robots are primarily composed of two types of physical light-sensing mechanisms with complementary strengths and weaknesses.  First, active sensing lidars to accurately represent metric space using point clouds. Secondly, passive sensing cameras for representing rich semantic information about the environment. Sensor fusion approaches aim at leveraging the complementary strengths of both vision modalities~\cite{hendy2020fishing-net}.

Semantic point clouds are the natural data structure for representing both spatial and semantic information.
Image content can be projected to a 3D point cloud if the pixel-wise depth and camera calibration parameters are known~\cite{forstner2016photogrammetric_cv}.
In principle, monocular~\cite{wang2019pseudolidar, guizilini2020sem_guided_depth, guizilini2020Packnet, godard2019digg_ss_mono_depth} or stereo vision~\cite{kendall2017e2e_stereo, khamis2018stereonet, chang2018psmn, xu2020aanet} can provide depth maps and enable a vision-only perception configuration equivalent to biological vision systems~\cite{medathati2016bio_vision}. However, we opted for a sensor fusion approach as current depth estimation methods result in excessively noisy estimates compared with lidar measurements.

Our framework processes observations as follows. First, the agent is instantiated within an unknown metric vector space.
Next, sensor observations are projected onto a common vector space by known intrinsic and extrinsic calibration parameters of each sensor with respect to the agent~\cite{geiger2013kitti}. For simplicity, we assume all sensor observations are synchronized into discrete timesteps.
We first infer semantics from images using a pre-trained semantic segmentation model~\cite{chen2018deeplabv3+} that partitions the image into distinct semantic regions. The 3D point cloud is projected onto the image frame resulting in a one-to-many mapping between semantic pixels and 3D points. The semantic information is appended to all respective points. Points without semantic information are discarded.

The set of all encounterable object semantics cannot be defined a priori in the open-world assumption. However, it is always possible to infer whether or not any new novel object possesses a known semantic.
Following this principle, we semantically partition the static environment into ``road'' and ``not-road'' observations.
Dynamic objects are not part of the static environment and should be ignored. For simplicity, we infer all dynamic objects as ``not-road'' observations and generally rely on temporal observation integration to filter out the resulting noise. The noise resulting from cars is problematic, as cars are large, abundant, and often parked on the road. For this reason, we infer ``car'' semantics as a special case for filtering out excessive noise. Alternatively, dynamic objects can be removed by applying a filtering method~\cite{carballo2018high, fu2022map_cleaner}.

The resulting representation is an agent-centric 3D semantic point cloud as visualized in the top row of Fig.~\ref{fig:semantic_point_cloud_representation}.


\subsection{Temporal observation accumulation}
The agent collects a sequence of temporally ordered semantic point clouds during operation. The goal is to integrate all observations into a single vector space centered on the agent. The least assumptions approach to estimate motion is by computing the transformation optimally matching two sequential observations by point cloud registration also known as scan matching~\cite{Thrun2005ProbRob}. We found that the iterative closest point (ICP) algorithm~\cite{besl1992icp} on unfiltered point clouds results in a simple but sufficiently accurate and robust implementation for accumulating observations covering $80 \times 80 \: m^2$ suburban road environments without revisitation. The agent trajectory is computed from the sequence of transformations.

ICP takes the previous and latest point cloud and computes the transformation aligning the previous point cloud $p^{(t)}$ to the latest one $p^{(t+1)}$. This transformation matrix $T_{t \rightarrow t+1}$ corresponds to the agent motion during the time difference between the two observations as shown in~(\ref{eq:icp_transformation_matrix}). We update the relative position of all previously accumulated point clouds $P^{(t)}$ to $\tilde{P}^{(t+1)}$ every time step recursively by applying the transformation as a matrix multiplication as in~(\ref{eq:icp_transformed_accum_obs}). Finally we add the new observations $p^{(t+1)}$ to the transformed accumulated observations $\tilde{P}^{(t+1)}$, resulting in a new set of accumulated observations $P^{(t+1)}$ as in~(\ref{eq:icp_new_accum_obs})

\begin{align}
    T_{t \rightarrow t+1} &= ICP(p^{(t)}, p^{(t+1)}) \label{eq:icp_transformation_matrix} \\
    \tilde{P}^{(t+1)} &= T_{t \rightarrow t+1} P^{(t)} \label{eq:icp_transformed_accum_obs} \\
    P^{(t+1)} &= concatenate(\tilde{P}^{(t+1)}, p^{(t+1)}) \label{eq:icp_new_accum_obs}.
\end{align}

Fig.~\ref{fig:observation_processing} show the agent trajectory from lidar odometry as a line. See Fig.~\ref{fig:semantic_point_cloud_representation} for a visual demonstration of accumulated semantic point clouds. 

\subsection{Partial world state representation}

Accumulated observations are projected onto homogeneous probabilistic grids representing world states. In contrast to 3D point clouds, 2D discrete grids can be processed by convolutional neural networks (CNN)~\cite{lecun1989zip-cnn} forming the backbone of recent potent generative models for images~\cite{Child21VDVAE, salimans2017pixelcnn++, ramesh2022dale2, saharia2022imagen}.

The projection is performed as follows.
First, we initialize a 2D grid $x^{(c)}$ spanning $(I, J)$ elements covering a rectangular spatial region with lengths $(H, W)$ for each semantic class or value $c \in (1, \ldots, C)$.  Next, the accumulated semantic point cloud $P$ is projected onto the grids.

We represent semantic information by beta distributions $p(x^{(c)}_{i,j} = True)$ modeling the Bayesian probability of element $(i,j)$ represent a semantic $c$. This formulation allows a single element to represent several semantics. The beta distribution is computed by counting the number of semantic points that confirm or refute the semantic within each grid element~\cite{mcelreath2020statistical-rethinking}.
Note that probability distributions allow representing ignorance or lack of knowledge. For example, if a grid element has no observations, the distribution $p(x^{(road)}_{i,j})$ is uniform, indicating unknown uncertainty.
Value information, like road surface lidar intensity measurements, are expressed as Gaussian distributions representing the mean and standard deviation of all observations encompassed by the grid element.
Finally, we concatenate all probabilistic and scalar 2D grids into a 3D tensor representation $x$. See Fig.~\ref{fig:observation_processing} and Fig.~\ref{fig:data_augmentation} for partial world state representation visualizations.

\section{Predictive world model}
The predictive world model samples diverse and plausible complete worlds conditioned on partially observed worlds as illustrated in Fig.~\ref{fig:front_figure}.
We implement the world model as an arbitrary conditioning generative model.
The model is trained by self-supervised learning to predict future observations from present observations akin to the predictive coding problem~\cite{lotter2017deep_pred_coding, marino2019pred_coding_vae_bio ,pearson2021multimod_pred_coding}.

Note that our method does not assume that integrating future observations necessarily results in complete observations. In early experiments we found that learning a model to predict complete world states from partially observed world states is not a trivial problem.
One challenge is conditioning by high-resolution dense partial observations. Another challenge is the lack of a complete ground truth learning signal, as typically learning to predict empty structure (i.e. predicting ``nothing'') is an easier solution than predicting plausible structure when lacking a target. Both issues rule out modeling by GANs~\cite{cai2020piigan, liu2021pdgan} but naturally lends themselves to VAEs~\cite{zheng2019picnet, zhao2020uctgan, peng2021dsi-vqvae}.

\begin{figure}
\centering
\includegraphics[width=0.42\textwidth]{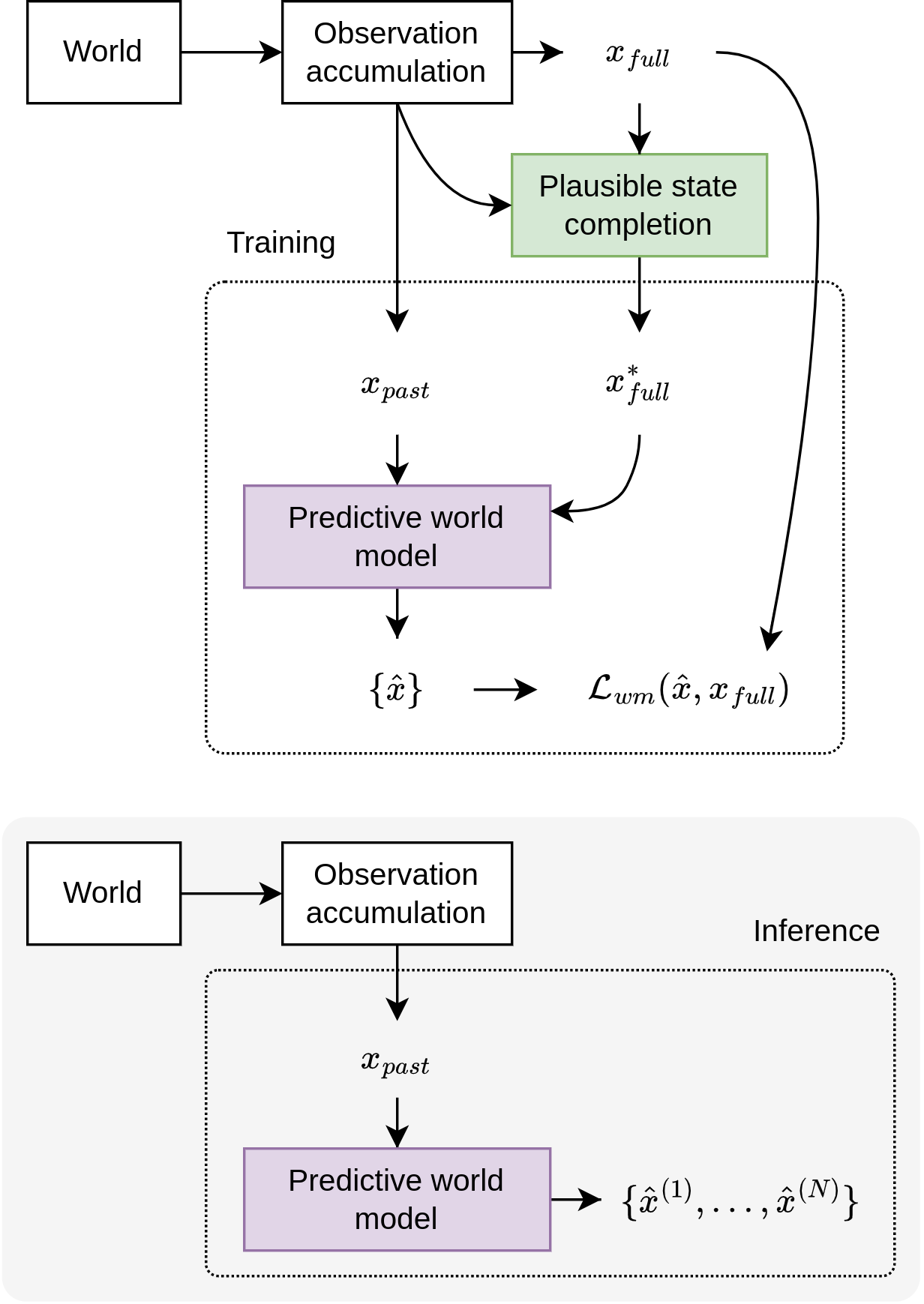} 
\caption{Overview of all modules in our framework. The goal is to predict complete worlds $\hat{x}$ conditioned on partially observed worlds $x_{past}$ (bottom). We train the predictive world model (purple) by first training an auxiliary module (green) providing pseudo ground-truth world states $x^*_{full}$ (top).}
\label{fig:model_overview}
\end{figure}

We present a novel solution extending the capability of hierarchical VAEs to learn to predict complete states from partial states only. Our method is formulated as a two-stage training process as illustrated in Fig.~\ref{fig:model_overview}.

\begin{figure}
\centering
\includegraphics[width=0.36\textwidth]{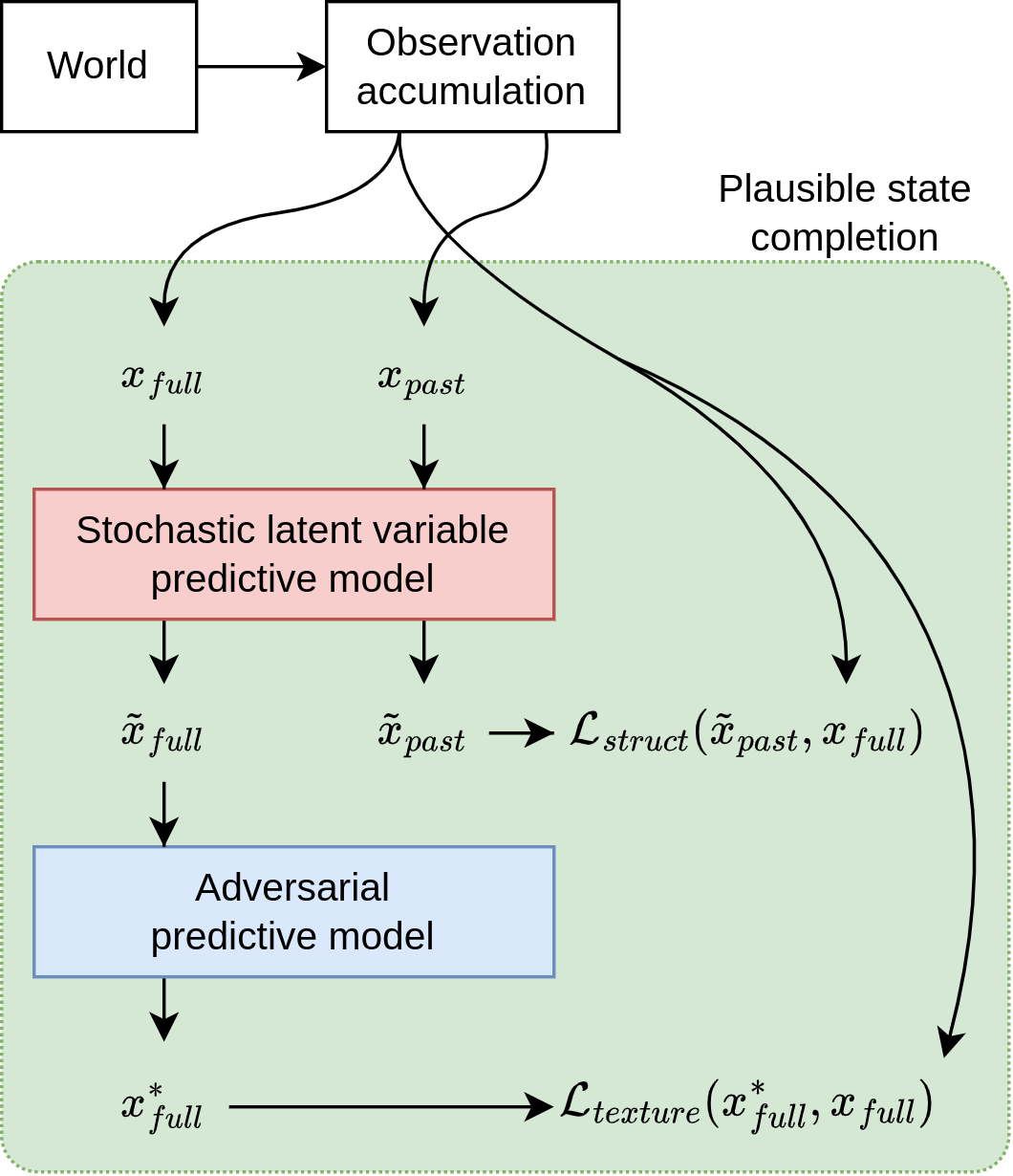} 
\caption{Overview of the plausible state completion module. The stochastic latent variable predictive model learns to predict missing structure $\tilde{x}_{past}$ from future observations $x_{full}$. The trained module is used to predict missing structures never observed, structurally completing the full observation $\tilde{x}_{full}$. The adversarial predictive model learns to fill in texture-like content in the predicted structure resulting in the pseudo ground-truth world state $x^*_{full}$.}
\label{fig:plausible_state_completion}
\end{figure}

In the first stage, we train two auxiliary models to generate a single complete plausible state by filling in the remaining unobserved elements after having integrated both past and future observations $x_{full}$. The first auxiliary model is a masked stochastic latent variable model that predicts structure (i.e. road region). The second auxiliary model is an adversarial model that generates texture (i.e. road surface intensity) for predicted structure elements. In the second stage, we use the complete plausible world states as pseudo ground truth states in order to train a more expressive HVAE capable of predicting complete states $\hat{x}$ from past observations $x_{past}$ only.

We implement the predictive world model by the recent SOTA hierarchical VAE model VDVAE by Child~\cite{Child21VDVAE}. The VDVAE model is capable of learning a rich hierarchical distribution of latent variables for high-resolution images, and achieves higher likelihoods than SOTA autoregressive models like PixelCNN~\cite{salimans2017pixelcnn++} while using fewer parameters and generates samples thousands of magnitudes quicker~\cite{Child21VDVAE}. The original non-hierarchical VAE model~\cite{kingma2013} learns a compact semantic latent code for images using a single stochastic layer. However, the reconstructed output tends to be blurry for high-resolution images. Adding stochastic layers of latent variables at different feature resolutions efficiently improves the expressiveness of the latent code by decoupling the representation of structure in lower-level layers and appearance in higher-level layers. The reconstructed output can be highly detailed also for high-resolution images.

%

%
%

\subsection{Plausible complete state prediction}
\label{sec:pseudo_gt_world_state_generation}

The pseudo ground truth states are generated by a sequential process illustrated in Fig.~\ref{fig:plausible_state_completion}. The first auxiliary model in the process takes the partially observed world state $x_{full}$ and predicts a new completed world state $\tilde{x}_{full}$ which includes the environment structure for unobserved regions. By structure we mean regions corresponding to navigable space and possessing semantic and scalar information. The second auxiliary model predicts a completed representation $x^*_{full}$ which includes texture-like content such as road intensity values for the newly predicted regions. In the rest of this section we explain each auxiliary model in detail. Note that this process does not substitute the predictive world model as the process leverages future observations not available at inference time.

We perform data augmentation on the original partially observed world representations when training the predictive world model. Augmentations include random rotation, translation, and warping operations applied identically on all tensor layers. Geometric data augmentation is essential for the predictive model to learn geometric invariance for top-down spatial representations~\cite{karlsson2020dsla}. Additionally, we perform a random sequence of sharpening, blurring, and value scaling operations to reduce overfitting to particular observed road intensity patterns. A set of training samples generated by augmenting a single sample is shown in Fig.~\ref{fig:data_augmentation}.

\begin{figure}
\centering
\includegraphics[width=0.50\textwidth]{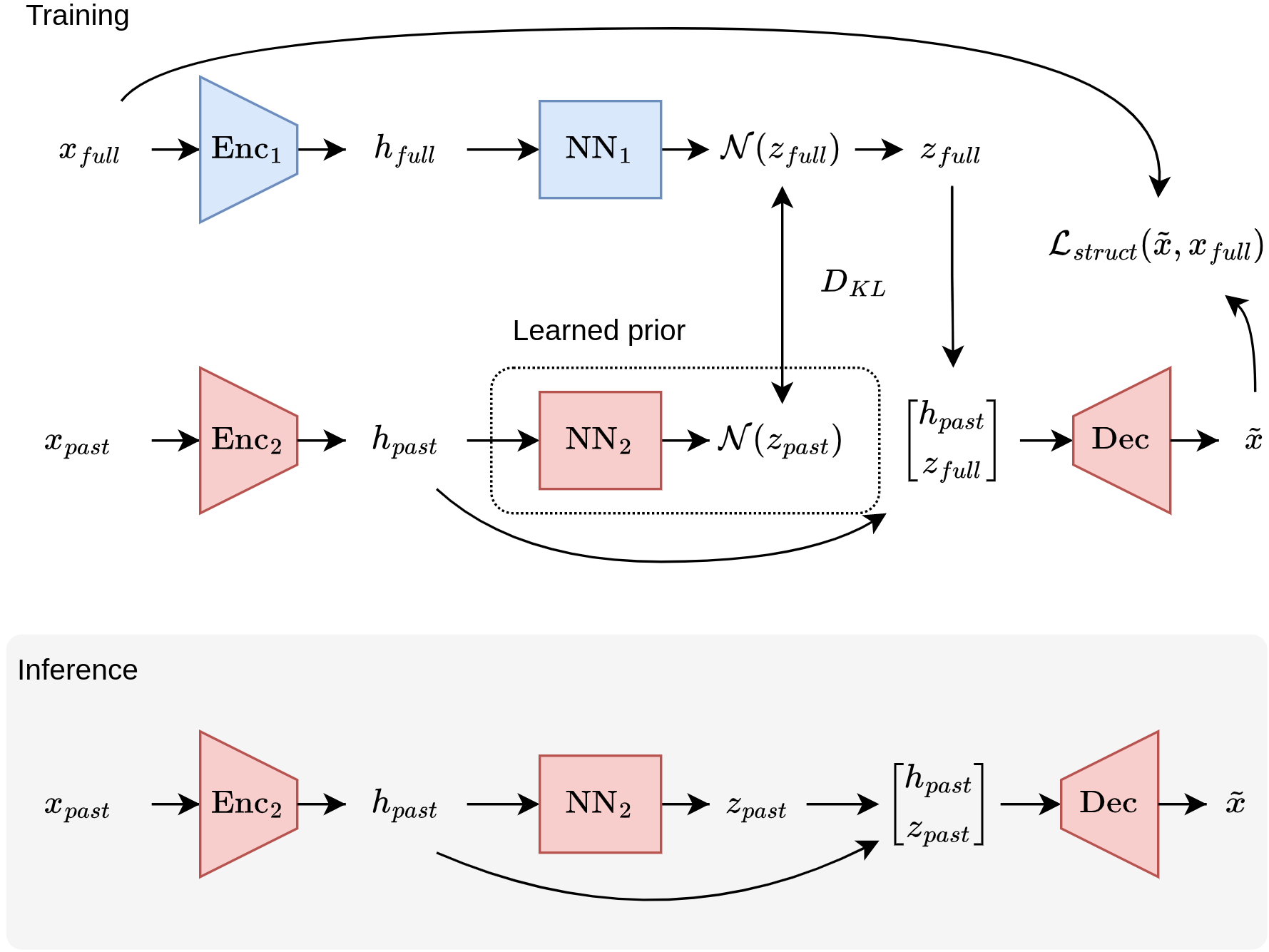}
\caption{Stochastic latent variable predictive model. An encoder-decoder is trained to predict the future observations $x_{full}$ from past observations $x_{past}$. A secondary encoder learns to encapsulate the inherent stochasticity by encoding the future observations as a latent code $z_{full}$. A learned prior is trained to predict the distribution for $z_{past}$ matching the distribution of $z_{full}$. Only the learned prior is used during inference.}
\label{fig:latent_variable_predictive_model}
\end{figure}

\begin{figure}
\centering
\includegraphics[width=0.48\textwidth]{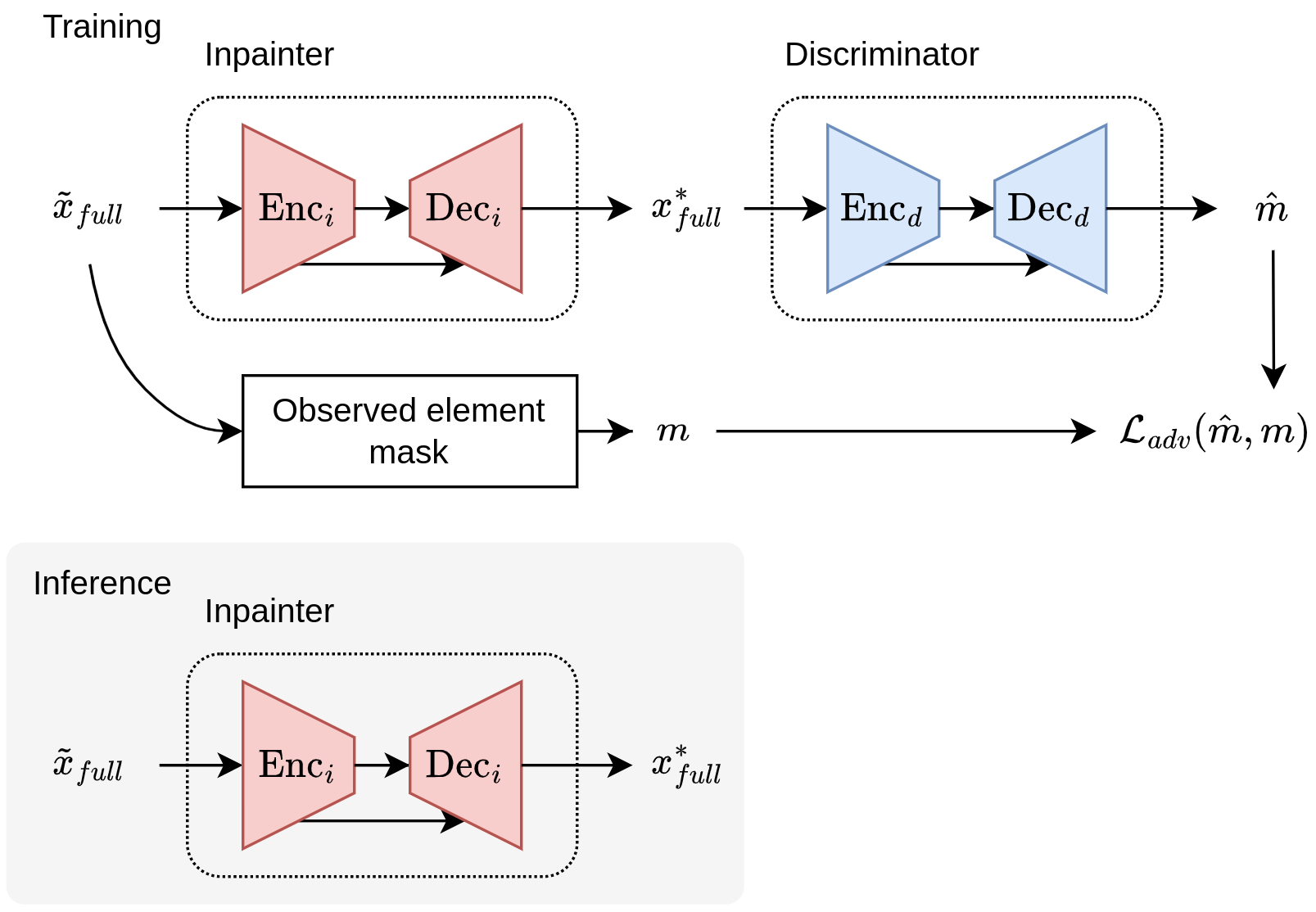}
\caption{Adversarial predictive model. An inpainting module is trained to fill in texture-like content indistinguishable from real observations. An element-wise discriminator module is trained to predict which regions are real and generated. Only the learned prior is used during inference.}
\label{fig:adversarial_predictive_model}
\end{figure}

\subsubsection{Predicting structure by a stochastic latent variable predictive model}
\label{sec:stochastic_latent_variable_predictive_model}
The model is structured as a dual path latent variable encoder-decoder model. When training the model, the first encoder takes the full state $x_{full}$ and predicts a latent variable distribution $Z_{full}$. The second encoder takes the past state $x_{past}$ and predicts a latent variable distribution $Z_{past}$. The distributions $Z_{full}$ and $Z_{past}$ are optimized to be similar. Next, a latent variable $z_{full}$ is sampled from $Z_{full}$ and appended to the encoding $h_{past}$ and feed to a decoder generating a new predicted complete state $\hat{x}$.
The model is optimized by computing the $\ell_2$ loss between $\hat{x}$ and observed elements in $x_{full}$

\begin{equation}
    \mathcal{L}_{struct} = \frac{1}{N_{struct}} M_{full} \odot ( \hat{x} - x_{full} )^2
\end{equation}

\noindent
where $N_{struct}$ is the number of structure elements, $M_{full}$ is a binary mask indicating observed elements in $x_{full}$ used to limit the loss to observed elements as is common in masked VAE methods~\cite{nazabal2018hi-vae, ma2019eddi, ma2020vaem, peis2022hh-vaem, collier2020vae_missing}.

The intuiting is that the learned distribution $Z_{full}$ contains the information required to reconstruct $x_{full}$, and the second encoder learns to estimate this distribution from $x_{past}$. In other words, knowing $Z_{full}$ explains away the stochasticity involved in reconstructing $x_{full}$ from $x_{past}$. At inference time $z_{full}$ is not known, but the second encoder has learned to predict $Z_{past}$ that is close to $Z_{full}$, effectively acting as a learned prior that explains away the stochasticity by a latent variable.

%
%
%
\subsubsection{Predicting texture by an adversarial predictive model}
Non-hierarchical reconstruction-based VAEs are ill-suited for generating fine-grained details~\cite{Child21VDVAE, vahdat2020nvae}. We therefore use an adversarial predictive model to generate pixel-like content such as road intensity for newly predicted structure. The model design is shown in Fig.~\ref{fig:adversarial_predictive_model}. The model consists of an encoder-decoder inpainting module that takes the world state $\tilde{x}_{full}$ with a completed structure, and generates a new world state $x^*_{full}$ also with completed content. The inpainting module is optimized by the minimax adversarial loss~\cite{pathak2016context_encoder, goodfellow2020gan} to make observed and generated content indistinguishable, while an element-wise discriminator module~\cite{schonfeld2020unetgan} is optimized to discriminate elements. The adversarial loss is computed  using the binary cross entropy (BCE) objective

\begin{equation}
    \mathcal{L}_{texture} = - \frac{1}{N_{struct}} \sum_{(i,j)} BCE(m_{(i,j)}, \hat{m}_{(i,j)})
\end{equation}

\noindent
where $m$ is the real observed element binary mask, $\hat{m}$ is the predicted mask by the discriminator, and $(i,j)$ are indices of structure elements. At inference time only the inpanting module is used to generate the completed world state $x^*_{full}$. 


\subsection{World model training}

\begin{figure}
\centering
\includegraphics[width=0.48\textwidth]{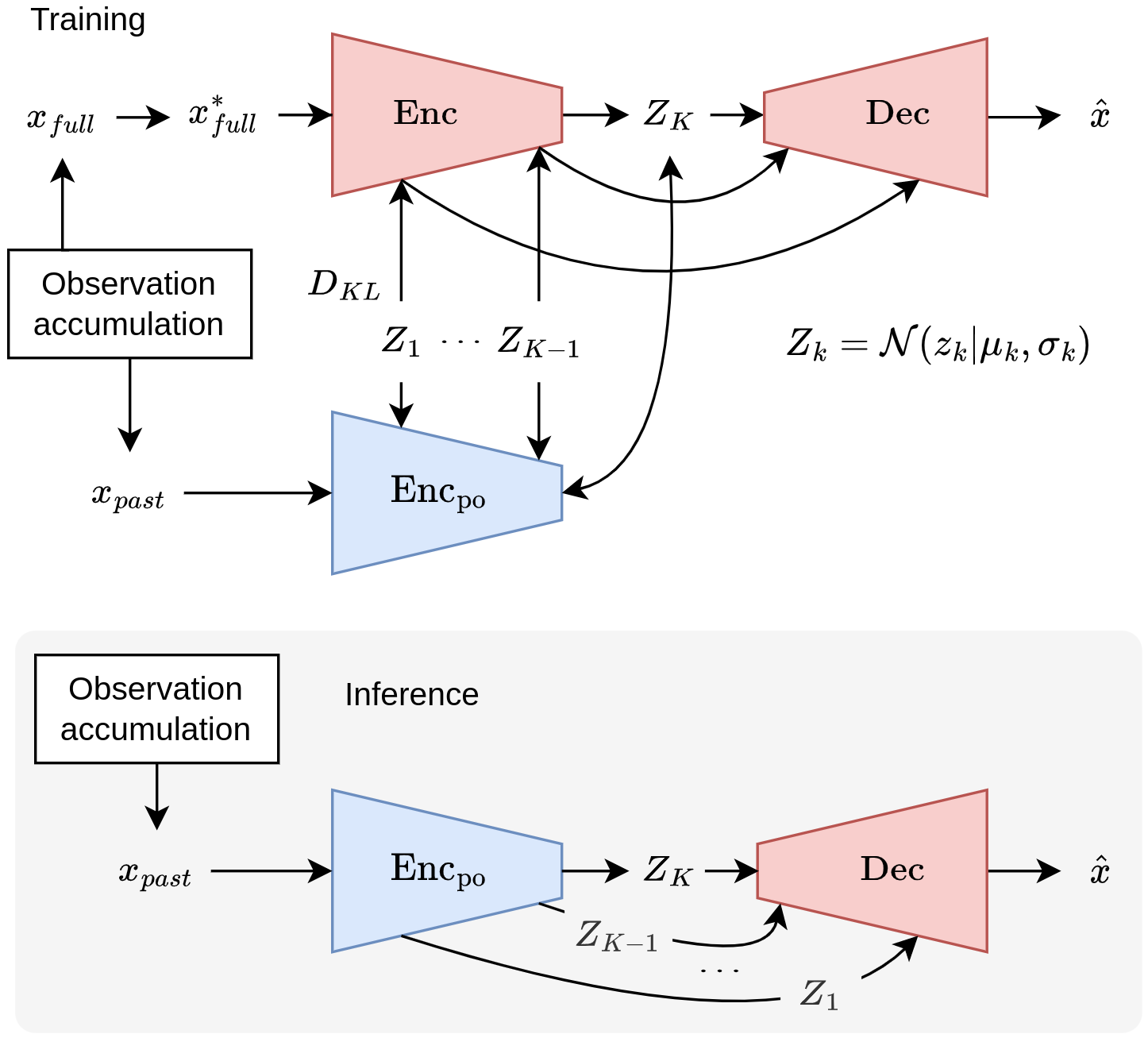}
\caption{Predictive world  model. We train a hierarchical latent variable generative model to reconstruct pseudo ground-truth world states $x^*_{full}$. Simultaneously, we train a secondary encoder to predict similar latent distributions from the original partially observed world states $x_{past}$ to predict complete world states $\hat{x}$ at inference time.}
\label{fig:generative_model}
\end{figure}

We train a model to predict a set of plausible worlds conditioned on partially observed worlds as depicted in Fig.~\ref{fig:generative_model}. First, we optimize a regular HVAE model~\cite{sonderby2016ladder_vae, ranganath2016hvae} parameterized by $q_\theta(z|x)$ and $p_\theta(x|z)$ to encode and reconstruct pseudo ground-truth world states $x^*_{full}$ generated by the plausible state completion module (see Sec.~\ref{sec:pseudo_gt_world_state_generation}). The learned hierarchical latent variable prior $p_\theta(z)$ and posterior $q_\theta(z|x)$ distributions~\cite{Child21VDVAE} can be factorized as

\begin{align}
 p_{\theta}(z) &= p_{\theta}(z_1 | z_2) \ldots p_{\theta}(z_{K-1} | z_K) p_{\theta}(z_K) \label{eq:hvae_prior} \\
 q_{\phi}(z|x) &= q_{\phi}(z_1 | z_2, x) \ldots q_{\phi}(z_{K-1} | z_K, x) q_{\phi}(z_K | x)
\end{align}

\noindent where each random variable is modeled by Normal distributions $\mathcal{N}(z|\mu, \sigma)$. Deeper or more abstract codes (i.e. $z_K$) encode the global structure, while shallow codes (i.e. $z_1$) encode the visual appearance of elements in $x^*_{full}$. We train the HVAE by maximizing the hierarchical ELBO

\begin{equation}
  \log p_\theta \ge \E_{q_\phi(z|x^*_{full})} \left[ \log p_\theta(x^*_{full}|z) \right] - D_{KL}(q_\theta(z|x^*_{full}) || p_\theta(z))
\end{equation}

\noindent where $\log p_\theta(x^*_{full}|z)$ is the likelihood of the reconstructed state, and a KL divergence term that measures the divergence between the distributions

\begin{equation}
\begin{gathered}
 D_{KL}(q_\theta(z|x^*_{full}) || p_\theta(z)) = \\
 \sum^K_{k=2} \E_{q_\theta(z_{\ge k}|x^*_{full})} \left[ D_{KL}( q_\theta(z_{k-1} | z_k, x^*_{full}) || p_\theta(z_{k-1} | z_k)) \right] \\
 + D_{KL} \left( q_\theta(z_K | x^*_{full}) || p_\theta(z_K) \right).
\end{gathered}
\end{equation}

We simultaneously train a secondary partially observed encoder $q_\phi(z|x_{past})$ to predict a latent distribution similar to $q_\theta(z|x^*_{full})$ based on the original partially observed world states $x_{past}$. The second encoder is optimized by minimizing


\begin{equation}
\begin{gathered}
 D_{KL}(q_\phi(z|x^*_{full}) || q_\psi(z|x_{past})) = \\
 \sum^K_{k=1} \E_{q(z_{>k}|x)} \left[ D_{KL}( q_\phi(z_k | z_{>k}, x^*_{full}) || q_\psi(z_k | z_{>k}, x_{past})) \right].
\end{gathered}
\end{equation}

At inference time the model uses the partially observed encoder to generate a latent distribution $q_\phi(z|x_{past})$ that can be decoded by $p_\theta(x|z)$ into a completely observed world state $\hat{x}$ similar to a pseudo ground-truth world state $x^*_{full}$ without the need to observe the future.


We developed the approach of optimizing hierarchical latent distribution similarity as an extension of the single layer latent variable model described in Sec.~\ref{sec:stochastic_latent_variable_predictive_model} inspired by the stochastic video prediction model by Denton~\cite{denton2018sto_vid_pred}. A similar approach named Posterior Matching by Strauss~\cite{Strauss2022PosteriorMatching} was very recently published concurrently to our work. Our method extends their method by allowing HVAEs to learn to predict complete states from partial states only. This property is important in continual learning real-world mobile robotics problems where the existence of a priori complete ground truth states cannot be presumed.

\section{Experiments}

We conduct three sets of experiments to verify the feasibility of each part of our framework in real-world environments. First, we evaluate the expected performance of pretrained vision-based semantic segmentation models used to interpret sensor observations. Secondly, we demonstrate the quality of generated partially observed world states. Finally, we evaluate the trained world model in terms of predictive accuracy and structural diversity.

%
Our self-supervised learning framework does not depend on human annotations and instead leverages a pretrained vision-based semantic segmentation models to infer semantic information from image observations. The first experiment set quantifies the expected domain generalization performance of models trained on one or several annotated public datasets and evaluated on our application target domain dataset. The training datasets are Apolloscape, BDD100K, Cityscapes, and Mapillary Vistas~\cite{wang2019apolloscape, yu2020bdd100k, cordts2016cityscapes, neuhold2017mapillary-vistas} providing 49287, 8000, 3475, and 20000 training samples, respectively. Our target domain dataset is KITTI-360~\cite{liao2021kitti360} providing 12054 annotated samples across all sequences.  All experiments are listed in Table~\ref{tab:semseg_domain_generalization}. All datasets provide Cityscapes-like labels and are thus easy to concatenate into a larger multi-domain dataset.
We use the SOTA semantic segmentation model DeepLabV3+~\cite{chen2018deeplabv3+} and evaluate performance using different ResNet backbones~\cite{he2015resnet} and number of training iterations. We leverage the MMSegmentation framework~\cite{mmseg2020} to train and evaluate models.

The second experiment set demonstrates how our framework generates BEV world state representations by integrating sequences of sensor observations. We found that leveraging 360 degree sensor observations result in a more useful temporal self-supervision learning signal than forward view observations only. As observations are not spatially and temporally biased in the driving direction, future observations are less obvious to predict, and thus improve extrapolation to all unobserved regions.
While KITTI-360 provides 360 degree vision from two fisheye cameras, it is not trivial to make use of these images. First, the availability of annotated fisheye image datasets is low. Secondly, camera calibration parameters for projecting the point cloud into the fisheye images are not specified.
We choose to qualitatively demonstrate the full perception pipeline on NuScenes~\cite{caesar2020nuscenes} as it provides 360 degree camera views and point clouds. We also quantitatively estimate semantic accuracy of the generated BEV representations on KITTI-360 by comparing the single forward-facing camera results and the corresponding ground truth point cloud semantics. We use the ``RN 101 320K cm'' model variant (see Table~\ref{tab:semseg_domain_generalization}) for segmenting images on both datasets.


\begin{table}
\caption{Semantic segmentation domain generalization performance.}
\begin{center}
\begin{tabular}{|c|c|c|c|c|c|c|}
  \hline
  Model  & Iters & Datasets$^{\ast}$ & mIoU  & road IoU & car IoU \\ \hline
  RN 18  & 80K   & ••cm     & 51.91 & 90.02    & 88.36   \\
         & 160K  & ••cm     & 53.52 & \textbf{92.30}    & \textbf{89.45}   \\
         & 320K  & ••cm     & \textbf{54.31} & 92.12    & 89.20   \\ \hline
  RN 50  & 80K   & •••m     & 55.30 & 92.84    & 89.90   \\
         &       & ••cm     & 55.78 & 89.71    & 90.30   \\
         &       & abcm     & 48.70 & 85.97    & 88.85   \\
         & 160K  & •••m     & 54.98 & 92.92    & 89.90   \\
         &       & ••cm     & \textbf{56.67} & 91.70    & \textbf{90.45}   \\
         &       & abcm     & 50.15 & 86.82    & 89.34   \\
         & 320K  & •••m     & 56.04 & \textbf{93.67}    & 89.43   \\
         &       & ••cm     & 56.65 & 93.56    & 90.26   \\
         &       & abcm     & 53.27 & 90.76    & 89.11   \\ \hline
  RN 101 & 80K   & •••m     & 56.96 & 93.44    & 90.23   \\
         &       & ••cm     & 56.00 & 93.23    & 89.74   \\
         &       & •bcm     & 54.98 & 93.12    & 89.27   \\
         &       & abcm     & 51.81 & 88.74    & 88.67   \\
         & 160K  & •••m     & 55.57 & 93.30    & 90.29   \\
         &       & ••cm     & 58.00 & 94.23    & 90.48   \\
         &       & •bcm     & 55.85 & 93.39    & 90.23   \\
         &       & abcm     & 51.71 & 90.20    & 89.55   \\
         & 320K  & ••cm     & \textbf{58.72} & \textbf{94.24}    & \textbf{90.55}   \\
   \hline
   \multicolumn{6}{l}{$^{\ast}$a: Apolloscape, b: BDD100K, c: Cityscapes, m: Mapillary V., •: Filler}
\end{tabular}
\label{tab:semseg_domain_generalization}
\end{center}
\end{table}

\begin{table}
\caption{BEV semantic segmentation performance on KITTI-360}
\begin{center}
\begin{tabular}{|c|c|}
    \hline
     & road IoU \\
    \hline
    Evaluate all regions & 92.31 \\
    \hline
    Evaluate unobserved regions only & 90.97 \\
    \hline
\end{tabular}
\label{tab:semseg_performance}
\end{center}
\end{table}

\begin{table}[H]
\caption{Mean and best world model prediction on the test sequence.}
\begin{center}
\begin{tabular}{|c|c|c|c|c|c|c|}
    \hline
      & \multicolumn{6}{|c|}{road IoU} \\
    \hline
    \#samples & 1 & 2 & 4 & 8 & 16 & 32 \\
    \hline
    Mean (all regions) & 97.94 & 97.94 & 97.96 & 97.94 & 97.94 & 97.94 \\
    \hline
    Best (all regions) & 97.94 & 98.18 & 98.38 & 98.52 & 98.63 & \textbf{98.73} \\
    \hline
    Mean (unob. only) & 96.21 & 96.22 & 96.23 & 96.21 & 96.21 & 96.21 \\
    \hline
    Best (unob. only) & 96.21 & 96.53 & 96.81 & 97.01 & 97.16 & \textbf{97.31} \\
    \hline
\end{tabular}
\label{tab:world_model_performance}
\end{center}
\end{table}

The third experiment set evaluates the performance of our predictive world model. We evaluate our model on the KITTI-360 dataset as it contains long driving sequences with high-frequency image and point cloud observations as expected in a real system. Long sequences improve learning to model large 80x80 m representations. High-frequency observations increase the element density of partially observed world states. We evaluate our model on sequence \#6 containing both suburban and urban road scenes. We use the remaining 8 sequences for training. We use the ground truth point cloud semantics in the predictive world modeling experiments due to KITTI-360 lacking usable 360 degree vision coverage.

\section{Results}


The experimental results validate our stated objectives in realistic real-world environments. The presented world model can predict accurate and diverse sets of plausible worlds through self-supervised learning and a pretrained vision-based semantic segmentation model.

\noindent
\textbf{Semantic segmentation performance} We present our experiment results in Table~\ref{tab:semseg_domain_generalization}. 
The consistently best training dataset combination is Cityscapes and Mapillary Vistas, 
and saturating the training data with a large set of regionally and camera-wise unsimilar samples, like Apolloscape, is detrimental for domain generalization performance on KITTI-360. Our hypothesis is that data regionally similar to the application domain, like Cityscapes, and diverse dataset capturing many regions and varying cameras, like Mapillary Vistas, are beneficial learning domain invariant features.
For backbone size we find that domain generalization performance improves with larger backbone sizes and additional training iterations.

\begin{figure}
\centering
\includegraphics[width=0.41\textwidth]{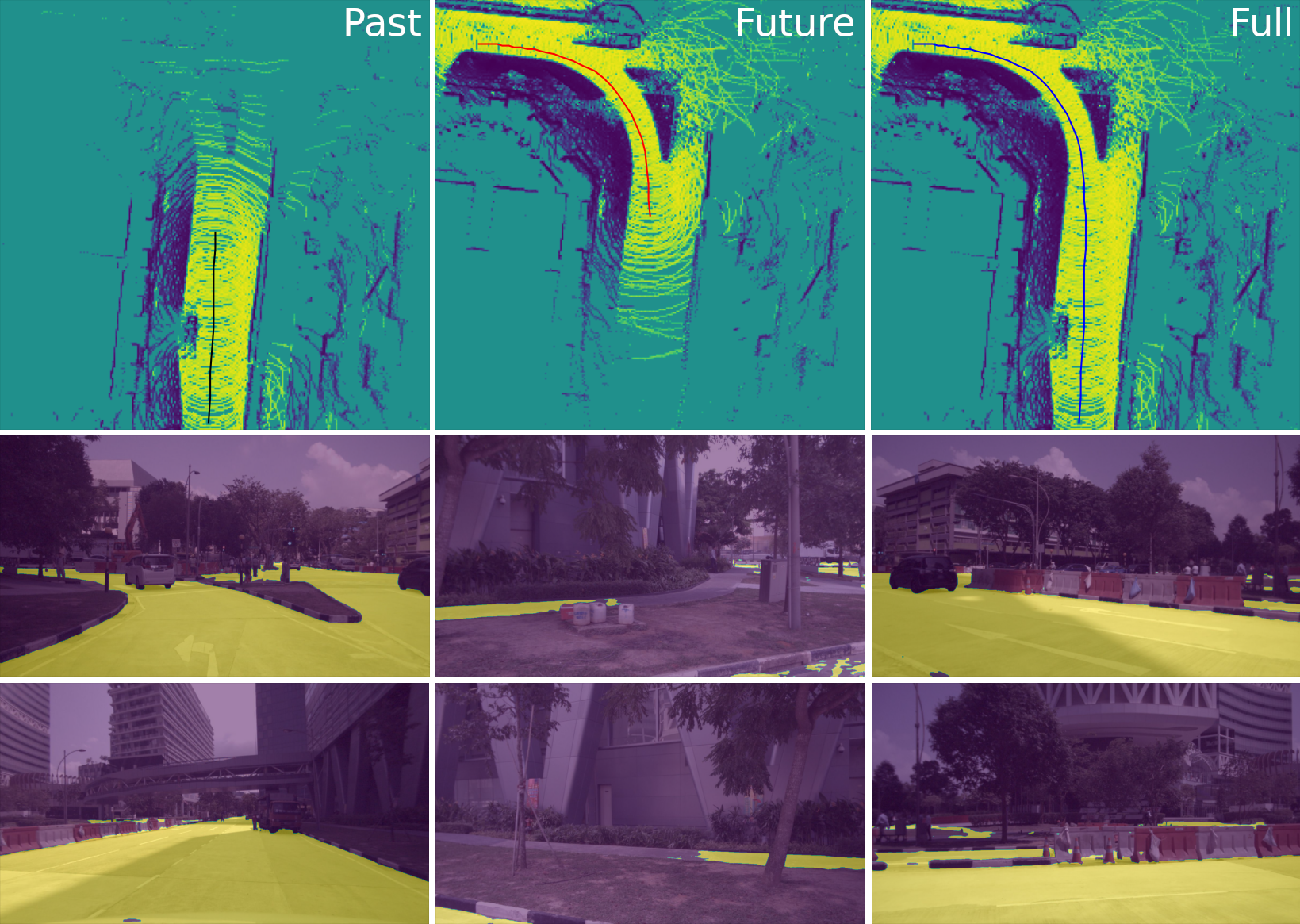} 
\caption{BEVs generated from our perception pipeline applied on NuScenes providing 360 degree camera setup and lidar. The top row show semantics. The bottom rows show semantic segmentation output.}
\label{fig:nuscenes_bev}
\end{figure}

\begin{figure}
\centering
\includegraphics[width=0.41\textwidth]{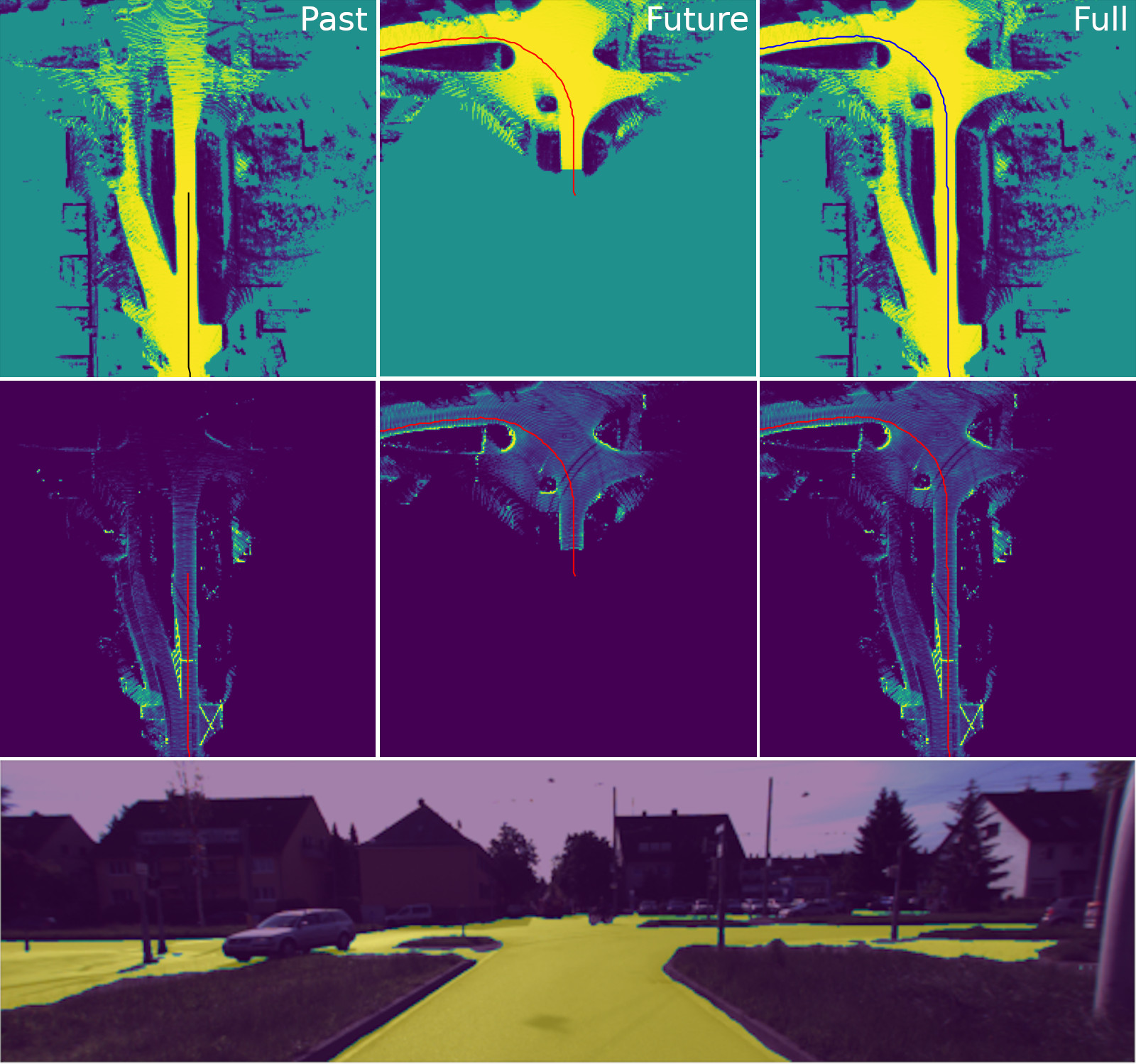} 
\caption{BEVs generated from our perception pipeline applied on KITTI-360 using one camera and lidar. The top row show semantics. The middle row show intensity. The bottom row shows semantic segmentation output.}
\label{fig:kitti360_bev_semseg}
\end{figure}

\begin{figure*}
\centering
\includegraphics[width=\textwidth]{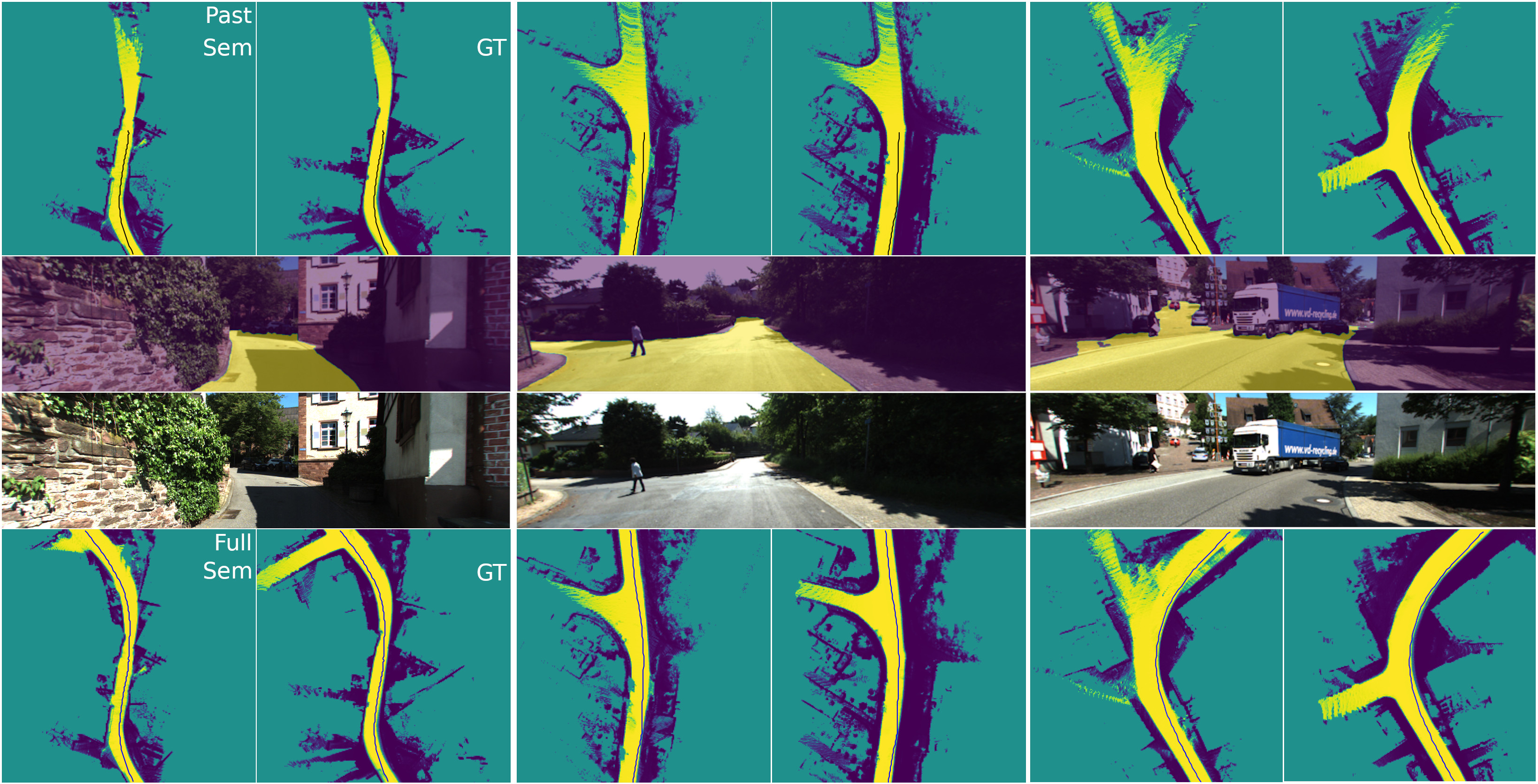}
\caption{The top row shows visual comparisons of accumulated observations generated using semantic segmentation or ground truth annotations. The middle row show predicted road semantics. The bottom row shows the fully integrated past and future observations. An example of the semantic ambiguity of human ``road'' annotation is shown in the rightmost example, where a seemingly road-like surface is not annotated as one, making perfect evaluation unattainable.}
\label{fig:semseg_gt_comparsion}
\end{figure*}

The high IoU values in Table~\ref{tab:semseg_domain_generalization} indicate that leveraging a pretrained semantic segmentation model is an adequate solution to infer relatively unambiguous semantic classes like ``road'' and ``car'' from image data.
See Fig.~\ref{fig:nuscenes_bev}-\ref{fig:semseg_gt_comparsion} for a visual demonstration of performance obtained on both NuScenes and KITTI~360.

Low IoU score samples are overrepresented by ambiguous classification of what is and is not ``road''. We conclude that the remaining performance gap is primarily a matter of semantic definition instead of a modeling problem. See Fig.~\ref{fig:semseg_gt_comparsion} for a visual demonstration of ambiguous semantics.

\noindent
\textbf{Generation and integration of semantic point clouds} We confirm that our framework is able to generate and temporally integrate semantic point clouds.
In Fig.~\ref{fig:nuscenes_bev} we show qualitative results on NuScenes based on the full 360 degree perception pipeline presented in Sec.~\ref{sec:generating_partial_world_states}.
Fig.~\ref{fig:kitti360_bev_semseg} shows results on KITTI-360 with a forward-facing camera only.
In Table~\ref{tab:semseg_performance} we present quantitative results comparing the ``road'' BEV generated by the pretrained semantic segmentation model and ground truth point cloud semantics. We present results evaluated for unobserved regions only (i.e. future) and also include previously observed regions (i.e. past + future). The results show that the BEV IoU score is comparable to that of the original perspective image IoU.

Our method lacks obvious comparative baselines. To the best of our knowledge, no prior work uses lidar observations and generative modeling to stochastically predict spatial environments without relying on ground truth map data. Prior image-based methods are generally non-generative models and are trained and evaluated on the same ground truth data domain. A reasonably fair comparison is between a recent SOTA image-based monocular model~\cite{yang2021pyva} reporting 68.34 road IoU on the full KITTI Raw dataset~\cite{geiger2012kitti}, and our model achieving 92.31 road IoU on the KITTI-360 dataset. Note that our results indicate the true expected domain generalization performance as the model is not trained on the same dataset domain, unlike the image-based model result. The degree of performance difference demonstrates the inherent advantage of using lidar observations for spatial prediction as presented by our method.

We observe an issue in open rural environments where incoming trucks may cause the ICP algorithm to lose point cloud correspondence. We believe a more robust probabilistic filtering approach~\cite{myronenko2009pcd, li2022robust_icp, zhang2022fast_robust_icp, vizzo2023kiss_icp} would remedy this problem.
Another issue is that very large structures, such as expressway intersections, are never comprehensively observed due to limited effective lidar observation range and thus difficult to learn to predict. We believe longer range lidars~\cite{carballo2020libre, lambert2020perf} and incorporating vision-based depth estimation methods~\cite{guizilini2020Packnet, xu2020aanet} may provide the necessary sensory range.

\noindent
\textbf{Predictive world model performance} Table~\ref{tab:world_model_performance} presents quantitative results showing the best IoU match among $N$ sampled complete world predictions and the actual future observed world. We find that in most situations future observations are deterministically predictable, meaning a single prediction gives a reasonable estimation of the true world. However, when the future observations are not deterministically predictable, sampling more worlds increases the likelihood of some prediction matching the future observations. 

\begin{figure*}[htbp]
\centering
\includegraphics[width=\textwidth]{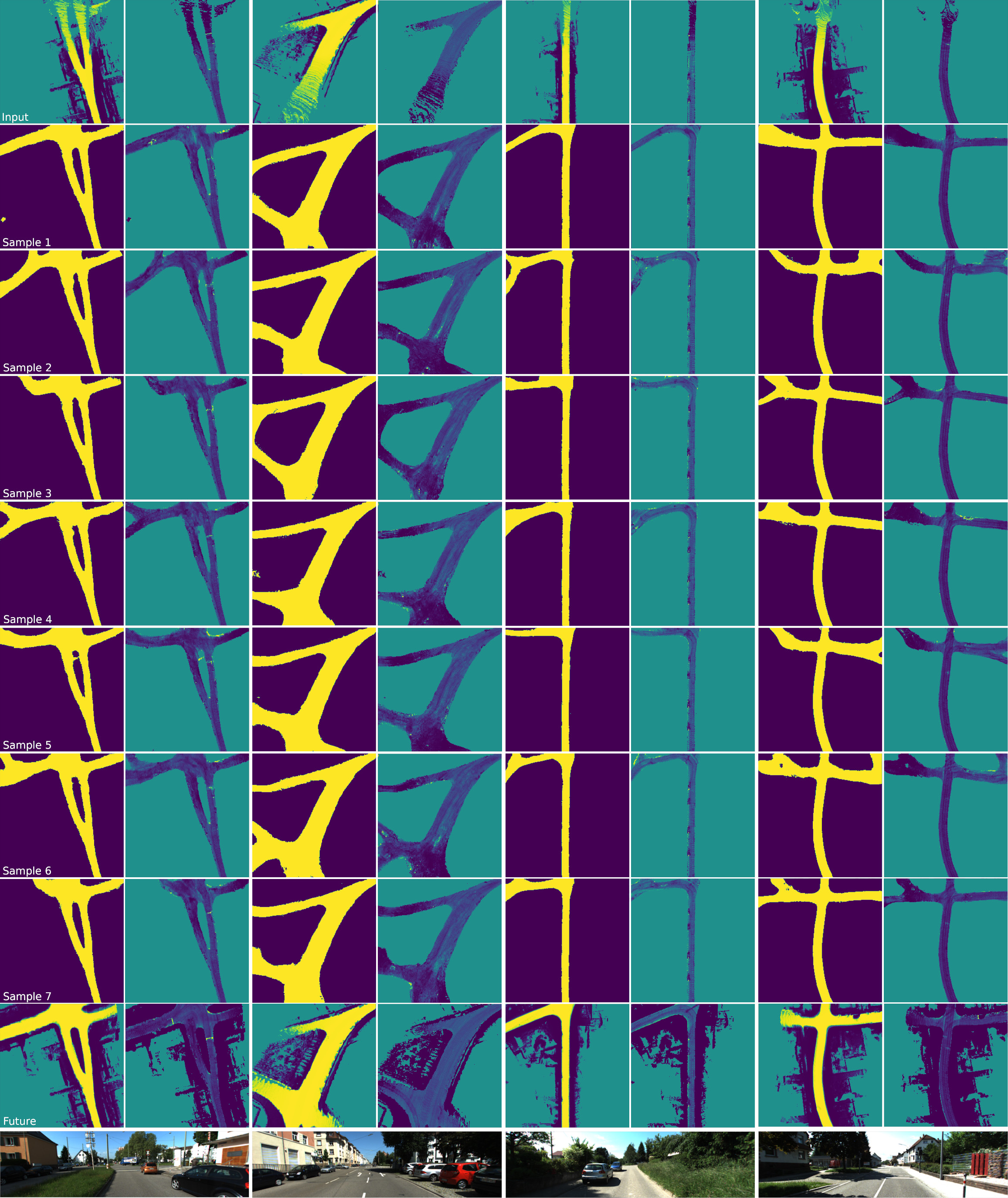}
\caption{The predictive world model can sample a diverse set of plausible complete worlds conditioned on a single partially observed world as input. The randomly sampled worlds demonstrate how the model can predict complex structures for unobserved regions of ambiguous road scenes. }
\label{fig:pred_model_sampling}
\end{figure*}

The relation between sampling and predictive performance is seen in Table~\ref{tab:world_model_performance} by how increasing the number of samples results in the best sample matching the future observations better (i.e. ``Best'') while the mean over all samples remains unchanged. The best prediction among 32 samples reaching 98.73 \% IoU, closing the gap to perfect prediction by 61.7~\% on average, when evaluating over both past and future observations.
See Fig.~\ref{fig:pred_model_sampling} for examples of sampled worlds.




\section{Discussion}
We present a framework to implement a self-supervised predictive world model for generating a diverse set of plausible complete world states trained from partially observed states only. The experimental results verify the feasibility of our approach in realistic real-world environments. We suggest the following improvements; a robust ICP implementation, longer range lidars for observing large road structures, and adding richer contextual semantics like RGB, lane markings, and traffic signs. Further work includes extending the world model to also represent and predict dynamic objects.

\section*{Acknowledgment}
This work was financially supported by JST SPRING, Grant Number JPMJSP2125. The authors would like to take this opportunity to thank the ``Interdisciplinary Frontier Next-Generation Researcher Program of the Tokai Higher Education and Research System''.

The computation was carried out through the ``General Projects'' program on the supercomputer ``Flow'' at the Information Technology Center, Nagoya University.


\end{document}